\pdfoutput=1

\documentclass[11pt]{article}

\usepackage{authblk}
\usepackage[]{acl}

\usepackage{times}
\usepackage{latexsym}

\usepackage[T1]{fontenc}

\usepackage[utf8]{inputenc}

\usepackage{microtype}

\usepackage{makecell}

\usepackage{url}

\usepackage{float}
\usepackage{graphicx}
\usepackage[caption=false]{subfig}

\usepackage{multirow}
\usepackage{enumitem}

\usepackage{amssymb}

\usepackage{color,soul}

\usepackage{todonotes}

\usepackage{footnote}
\usepackage{booktabs}

\definecolor{brightmaroon}{rgb}{0.76, 0.23, 0.28}

\title{The Power of Summary-Source Alignments}

\author[ \; 1,2]{\bf Ori Ernst\thanks{\;\; Work was done as an intern at Amazon.}}
\author[2]{\bf Ori Shapira}
\author[1,2]{\bf Aviv Slobodkin}
\author[2]{\bf Sharon Adar}
\author[2,3]{\\ \bf Mohit Bansal}
\author[1]{\bf Jacob Goldberger}
\author[2]{\bf Ran Levy}
\author[1]{\bf Ido Dagan}
{
\makeatletter
\renewcommand\AB@affilsepx{~~~~~~ \protect\Affilfont} \makeatother
\affil[1]{Bar-Ilan University}
\affil[2]{Amazon}
\affil[3]{UNC Chapel Hill}
}
\affil[  ]{} 
\affil[  ]{\tt \{oriern, obspp18, lovodkin93\}@gmail.com}
\affil[  ]{\tt \{sharonaz,ranlevy\}@amazon.com}
\affil[  ]{\tt \{mbansal\}@cs.unc.edu}
\affil[  ]{\tt \{jacob.goldberger@, dagan@cs.\}biu.ac.il}

\date{}

\begin{document}
\maketitle

\begin{abstract}

Multi-document summarization (MDS) is a challenging task, often decomposed to subtasks of salience and redundancy detection, followed by text generation.
In this context, alignment of corresponding sentences between a reference summary and its source documents has been leveraged to generate training data for some of the component tasks. Yet, this enabling alignment step has usually been applied heuristically on the sentence level on a limited number of subtasks.
In this paper, we propose extending the summary-source alignment framework by (1) applying it at the more fine-grained proposition span level, (2) annotating alignment manually in a multi-document setup, and (3) revealing the great potential of summary-source alignments to yield several datasets for at least six different tasks. Specifically, for each of the tasks, we release a manually annotated test set that was derived automatically from the alignment annotation. We also release development and train sets in the same way, but from automatically derived alignments.
Using the datasets, each task is demonstrated with baseline models and corresponding evaluation metrics to spur future research on this broad challenge.

\end{abstract}

\section{Introduction}
\label{sec_intro}
\begin{figure}[t]
    \centering
    \resizebox{0.8\linewidth}{!}{
    \includegraphics{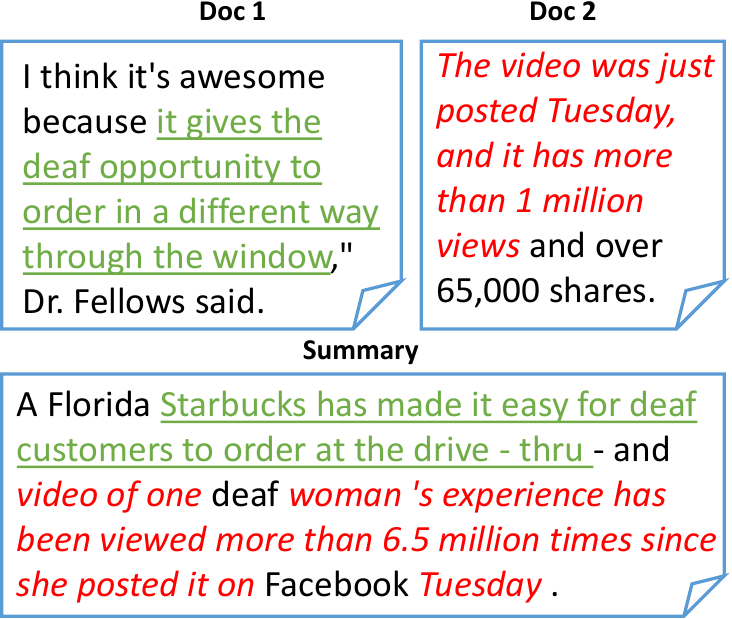}}
    \caption{An example of proposition-level multi-document-based alignment. Aligned propositions are in the same color and formatting.}
\label{fig:alignment_example}
\end{figure}

Common information needs are most often satisfied by multiple texts rather than by a single text.
Processing multiple texts is a challenging feat due to the wealth of content they possess, as well as the anticipated redundancy of information. To address this challenge, various tasks support multi-text processing needs, such as information selection, consolidation and fusion. A prominent application that aims to respond to user information needs is the multi-document summarization (MDS) task. Inherently, MDS either explicitly or implicitly prescribes sub-tasks like those listed above \citep{moryossef2019stepbystep, ernst2022procluster}.

Notably, many works showed the superiority of decomposing a complex task, and more specifically summarization, into its natural subtasks \citep{moryossef2019stepbystep, ernst2022procluster, zhang-etal-2023-enhancing, xiao2023salienceExplanationMDS}. However, in most cases there is no training data for each of the subtasks, nor gold test data to evaluate them. That is, while summarization datasets contain gold reference summaries, they do not publish, e.g., gold document salient spans for the \textit{salience detection} task, or gold sentences that fuse information from a few salient document spans for the \textit{fusion} task.

In this paper, we unveil a simple approach to obtain high-quality datasets for a wide variety of multi-text related tasks, via a single annotation process of summary-source alignments. Specifically, we match summary propositions with their proposition-level supporting evidence in the source over an existing MDS dataset \citep[Multi-News; ][]{fabbri-etal-2019-multi}. An example for alignments is presented in Figure \ref{fig:alignment_example}.

Aligning all the information segments between a reference summary and its paired document set reveals the underlying sub-tasks constructing the summarization process, as illustrated in \autoref{fig_derived_datasets}. For instance, the aligned spans in the document set constitute the salient information within them, since they collectively embody the corresponding summary. This characteristic captures a salience detection task (\autoref{fig_derived_datasets} (b)). Similarly, a summary segment can be viewed as a fused version of all its aligned document-set mentions, representing a sentence fusion task (\autoref{fig_derived_datasets} (f)).

Overall, with these alignments we automatically derive datasets for six tasks: (1) salience detection, (2) proposition coreference clustering, (3) evidence detection, (4) text planning, (5) sentence fusion, and (6) in-context passage fusion. In essence, this procedure ``reverse engineers'' the human summarization process by which the reference summaries were originally created. The resulting data enriches the current inventory of datasets for these individual tasks, while being derived automatically solely from the high-quality alignment annotations.
While the tasks addressed here stand on their own merits, such tasks have also been shown to benefit the overall summarization process when used within a pipeline, as mentioned before (more on this in \S{\ref{sec_background}}).

Our high-quality alignments test set was obtained through a controlled crowdsourcing procedure \citep{roit2020controlled}, with annotators diligently trained for this task, and contains 100 topics (document-set/summary pairs) with 2256 alignments.
We also created large-scale training and development sets by extracting alignments automatically, using the SuperPAL alignment model \citep{ernst2021superpal}, from the Multi-News train and dev sets (\S{\ref{sec_dataCollection}}). We automatically derive and release train and test datasets from the alignment data for the six mentioned tasks (\S{\ref{sec_tasks}}). For each of the tasks, and using its respective dataset, we develop and evaluate two baseline models explicitly targeting the task: one is a trained model while the other is a non-finetuned execution of a ChatGPT LLM \citep{openai2023api}
(\S{\ref{sec_benchmark}}). Over the six tasks, we generally find that smaller trained models yield better results than the GPT counterpart, leaving room for future advances using our task and dataset suite.

Overall, this work showcases that alignments from an MDS dataset empower a rich collection of multi-document related tasks. These tasks are appealing on their own, and are additionally advantageous as sub-components of MDS solutions.\footnote{All data is publicly available at \href{https://github.com/oriern/SPARK}{https://github.com/oriern/SPARK}.}

\section{Background}
\label{sec_background}
Aligning information between source and reference texts has been previously addressed in the realm of summarization. An early effort was conducted for the purpose of summary evaluation through the Pyramid method \citep{nenkova2004pyramid}.
It's effectiveness, albeit burden of annotation, triggered the pursuit of automatic
procedures that mimic the content extraction and alignment \citep{Yang2016peak, gao2018pyreval, hirao2018pyrEDU, zhang2021lite2pyr}, generally via proposition extraction and matching. 
The alignment approach and manual annotation process for our dataset is reminiscent of the Pyramid method's, however it is more scalable thanks to controlled crowdsourcing \citep{roit2020controlled}.
While \citet{shapira2019litepyr} also applied crowdsourcing for easing manual alignment, the summary is not exhaustively aligned and high quality is not guaranteed.

Besides summary evaluation, alignments have also been useful for the summarization process itself.
To acquire such alignments, the ROUGE metric \citep{lin2004rouge} was leveraged to match between summary and source sentences \citep{zhang2018neuralExtSumm, cho2019dpp}. 
This approach was also taken for components of the summarization pipeline, such as detection of salient sentences \citep{chen2018fastabs} and sentence fusion \citep{lebanoff2019scoreingpairs}.
The heuristic nature of this pairing approach yields noisy alignments since it is both on the sentence level and based on lexical matching.

Many additional works have shown the benefit of decomposed summarization pipelines. \citet{moryossef2019stepbystep} and \citet{ernst2022procluster} split the summarization task to planning and realization phases (corresponding to our derived tasks) showing improved summary outputs. \citet{zhang-etal-2023-enhancing} find that decomposition of the summarization process can improve faithfulness of the output summary to its source documents. \citet{xiao2023salienceExplanationMDS} shows the benefit of salience detection as an initial phase, both for improving summary quality, and to provide attribution to summary segments.

Information alignment has also been treated as a standalone task. \citet{ernst2021superpal} designed a supervised model that far exceeds the abilities of lexical-based aligners. They also released a high quality test set of sub-sentence-level alignments. 
Collecting data required expert cleaning of crowdsourced annotations. 
Our more scalable approach
yielded a dataset that is an order of magnitude larger (11 vs 100 topics), 
and further improves alignment accuracy through refined guidelines and removed UI constraints in the annotation tool \citep{slobodkin2022ctr}.

Recently, \citet{krishna2023usb} released USB, a benchmark for summarization-related tasks, also derived from alignments. Alignment was conducted between the leading section of a Wikipedia article and its body, via controlled crowdsourcing, and required editing text spans in the leading section to remain faithful to the article body.
In contrast to their \textit{sentence} level alignments, our \textit{proposition}-level  alignments eliminate non-aligning noise. Moreover, aligning in the \textit{multi}-document setting, as opposed the \textit{single}-document in USB, introduces challenges arising from cross-document information sharing and size. The differences listed above induce tasks in our work that are mostly different from those addressed in USB, and that can be treated as standalone tasks.

\section{Collecting Alignments Data}
\label{sec_dataCollection}

Our alignments data consists of a high quality test set, collected through careful manual annotation (\S{\ref{sec_dataCollection_test}}), as well as large-scale training and development sets that were automatically compiled (\S{\ref{sec_dataCollection_auto}}). All alignments were extracted from the respective data split of Multi-News \citep{fabbri-etal-2019-multi}, a MDS dataset of sets of news articles with professionally prepared summaries. 

An instance in our alignment dataset is based upon a document set $D$ and a corresponding reference summary $s$. The instance consists of a list of aligned pairs $H = \{(h^s_1, h^D_1), (h^s_2, h^D_2), ..., (h^s_n, h^D_n)\}$ such that $(h^s_i, h^D_i)$ are proposition spans from the summary and document set, respectively, that describe the same piece of information. Since information is expected to repeat across the documents, $H$ likely contains pairs where $h^s_i = h^s_j$. Moreover, the summary should be exhaustively covered, i.e., all propositions in $s$ are expected to appear in $H$.

\begin{table}[!t]
    \resizebox{\linewidth}{!}{
    \begin{tabular}{lrrr}
    \toprule
    \textbf{Stat} & \textbf{Train} & \textbf{Dev} & \textbf{Test} \\
    \midrule
    \# topics & 44.5K & 5567 & 100 \\
    \# alignments & 1.5M & 186K & 2256  	 \\
    \# clusters & 629K & 77.5K & 1332	 \\
    \# summary sentences & 342K & 42K &  834  \\
    avg. cluster size & 2.4 & 2.4 & 1.7    \\
    avg. \# clusters per sent & 1.8 & 1.8 & 1.6 \\
    avg. \# clusters per topic & 14.1 & 14.1 & 13.6 \\
    avg. \# docs per topic & 2.71 & 2.65 & 2.97 \\
    \bottomrule
    \end{tabular}}
    \caption{Statistics of our alignments data. The test set is manually collected and the dev/train sets are automatically collected, all from the Multi-News MDS dataset. A cluster contains alignments pertaining to document spans that align to the same summary span (referring to the same information).}
    \label{tab_dataStat}
\end{table}






         

     
     

\subsection{Manually Annotated Test Set}
\label{sec_dataCollection_test}

To manually collect the alignments from a document-set/summary pair (``topic''), we follow the annotation protocol of \citet{slobodkin2022ctr}, using controlled crowdsourcing \citep{roit2020controlled},\footnote{Potential annotators were first picked through a filtering crowdsourcing task, and then went through several increasingly challenging stages of alignment annotations for quality assessment and training on the task. Eventually five annotators were qualified and completed the tasks.} adapting their method to the multi-document setting and altering the annotation guidelines for our purposes. Our annotation yields 2256 alignments from 100 topics. Full statistics are in \autoref{tab_dataStat}.

\paragraph{Annotation interface and procedure.}
We adopt the web-based annotation tool from \citet{slobodkin2022ctr}, and deploy it on Mechanical Turk\footnote{\url{www.mturk.com}} for crowdsourcing (see \autoref{fig:UI_example} in Appendix). 
The tool shows documents and the corresponding summary side-by-side, and annotators identify matching text segments between a document and the summary. The annotator is instructed to concentrate on an individual summary statement at a time, and to eventually cover the full summary. Each topic is annotated by a single trained annotator. For quality assurance, submissions were randomly checked and direct feedback was given as needed.

.

\paragraph{Annotation guidelines.}
\label{subsec:guidelines}
A proposition in a summary is determined by a central event (predicate) with its associated arguments. Annotators are guided on how to identify these propositions, including special cases of nested propositions (a proposition being an argument of another), predicates connected via discourse markers, prospective vs. transpired events, and incontiguous propositions. See Appendix \ref{appendix_annotation_guidelines} for explanations and details.

Upon identifying an event in the summary, the annotator locates all aligning spans in the document. A span is defined as the minimal token-set that fully covers the summary proposition without including additional information. A document span may explicitly refer to the summary event, or entail it (the summary event might generalize several instances of coreferring document events).

\paragraph{Inter-annotator agreement.}
To assess the quality of alignments we measured inter-annotator agreement on a set of instances annotated by all five annotators. A total of 31 summary sentences were annotated against their respective document sets. For every pair of annotators we computed intersection-over-union (\textit{IoU}) of the token indices (only content words) in the document spans that align to the same summary sentence, akin to \citet{ernst2021superpal}. Over the 310 compared pairs (31 sentences with 5 workers), the resulting \textit{IoU} score is 0.717, suggesting high-quality annotation.

\subsection{Automatically Collected Data}
\label{sec_dataCollection_auto}

For training and development sets, we extracted alignments automatically, using SuperPAL \citep{ernst2021superpal}, from the Multi-News train and dev sets, respectively. Document propositions clustered to the same summary proposition were then clustered together. The train/dev sets have 1.5M/186K alignments from 44.5K/5.5K topics (see \autoref{tab_dataStat}).

\section{The Task Suite}
\label{sec_tasks}
\begin{figure*}[t]
\centering
\subfloat[Alignments]{\includegraphics[scale=.2]{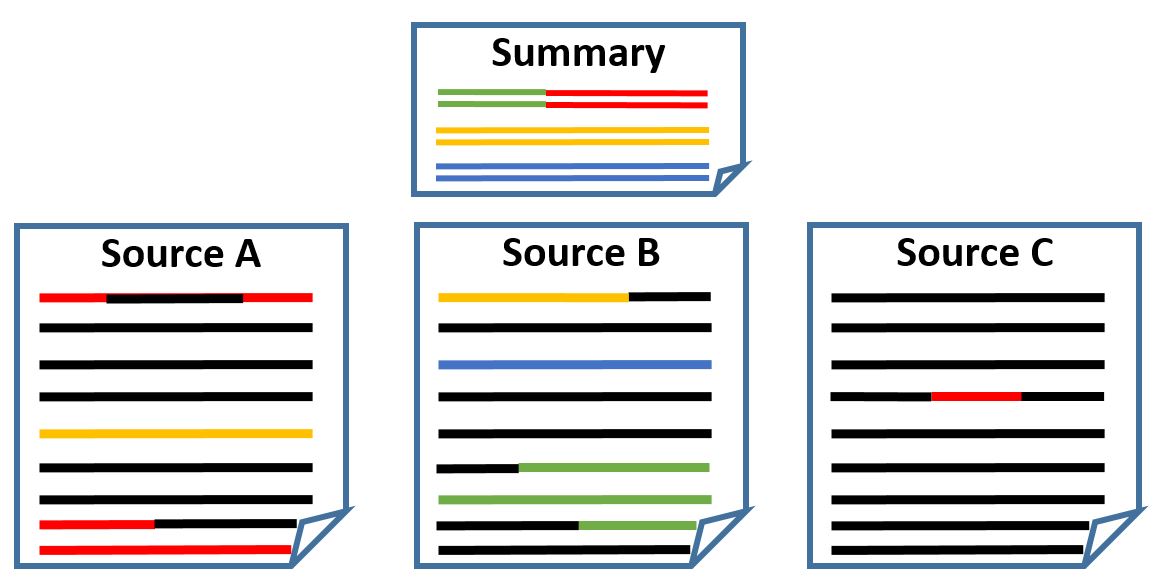}}\label{fig_derived_datasets_alignment}\\
\vspace{-.4cm}
\subfloat[Salience Detection]{\includegraphics[scale=.17]{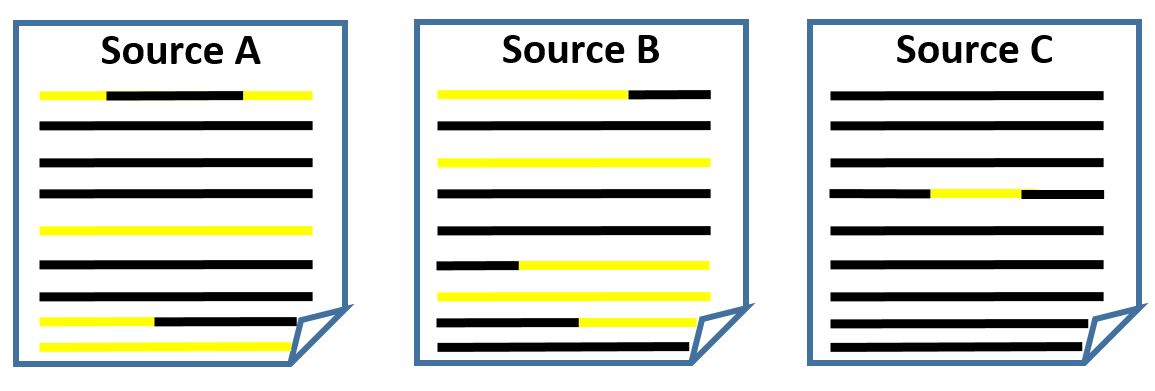}\label{fig_derived_highlighting}}
\hspace{.15cm}
\subfloat[Proposition Clustering]{\includegraphics[scale=.13]{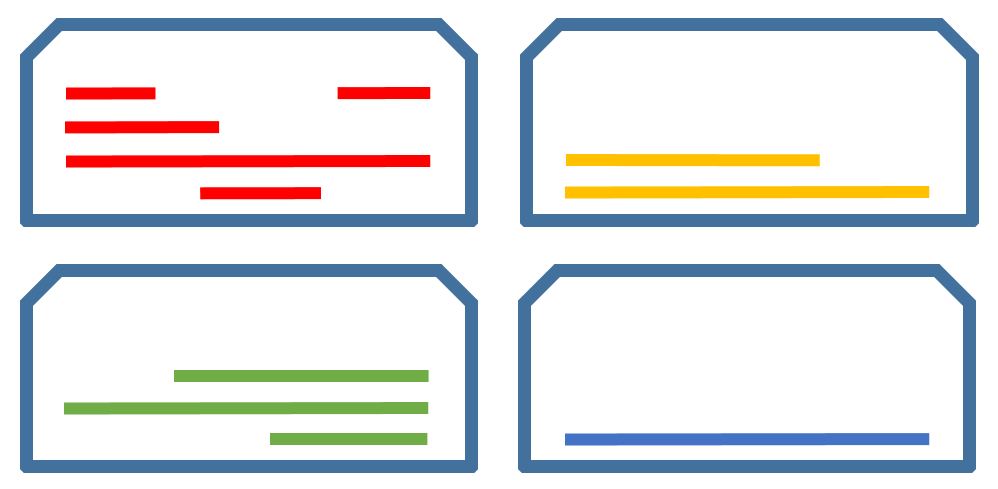}\label{fig_derived_clustering}}
\hspace{.15cm}
\subfloat[Evidence Detection]{\includegraphics[scale=.17]{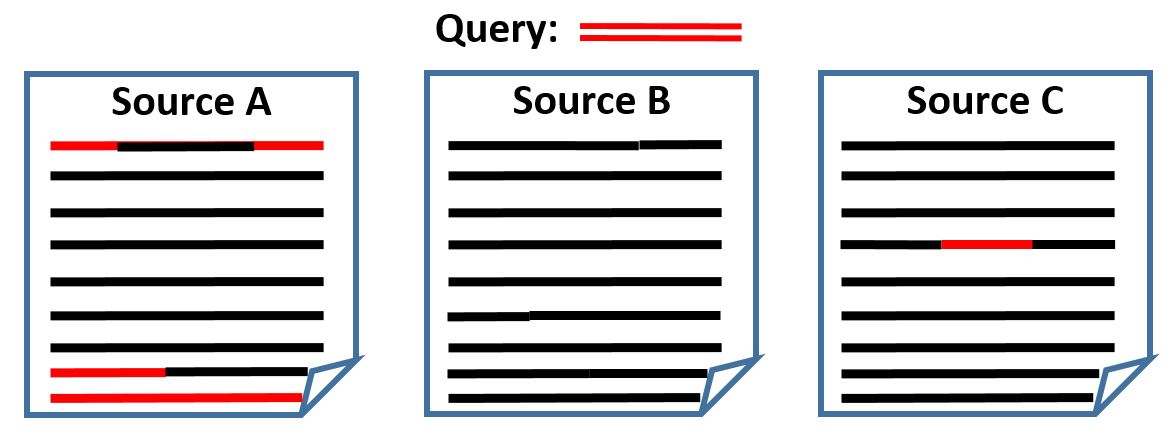}\label{fig_derived_evidence}} 
\hspace{.15cm}
\subfloat[Sentence+Paragraph Planning]{\includegraphics[scale=.185]{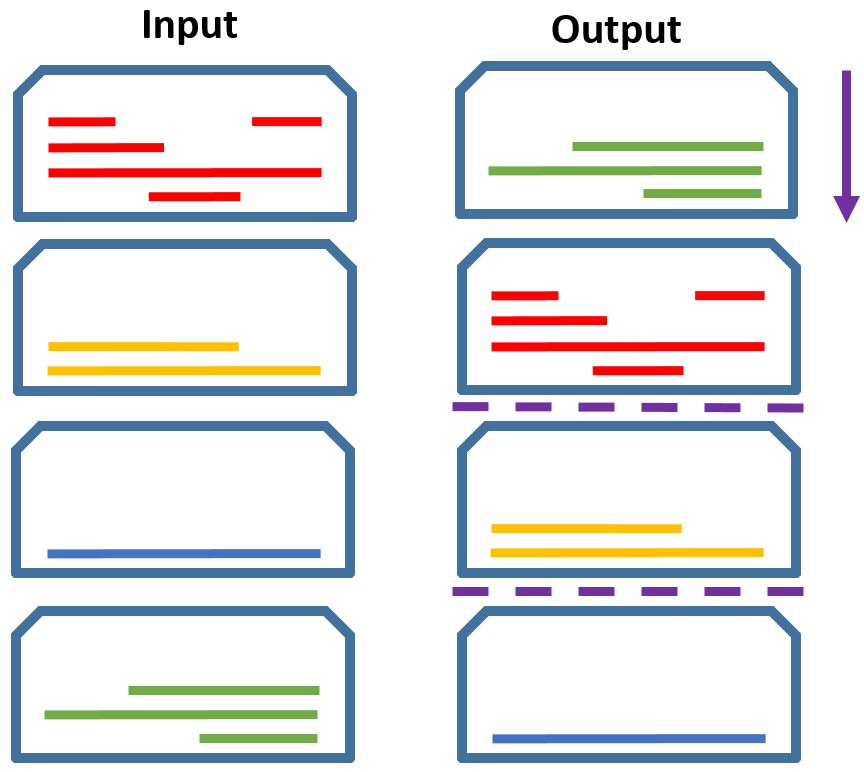}\label{fig_derived_planning}} 
\hspace{.20cm}
\subfloat[Sentence Fusion]{\includegraphics[scale=.185]{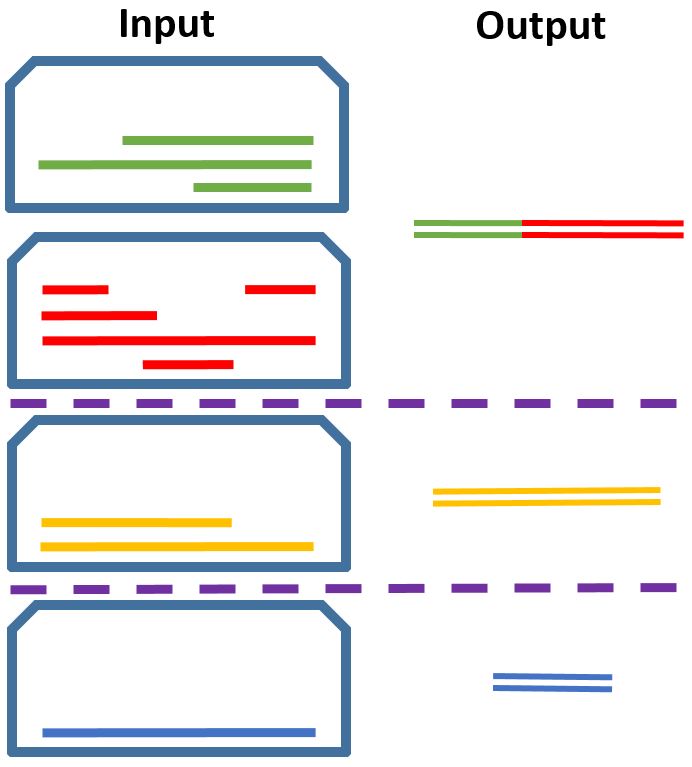}\label{fig_derived_datasets_fusion}} 
\hspace{.18cm}
\subfloat[In-context Passage Fusion]{\includegraphics[scale=.20]{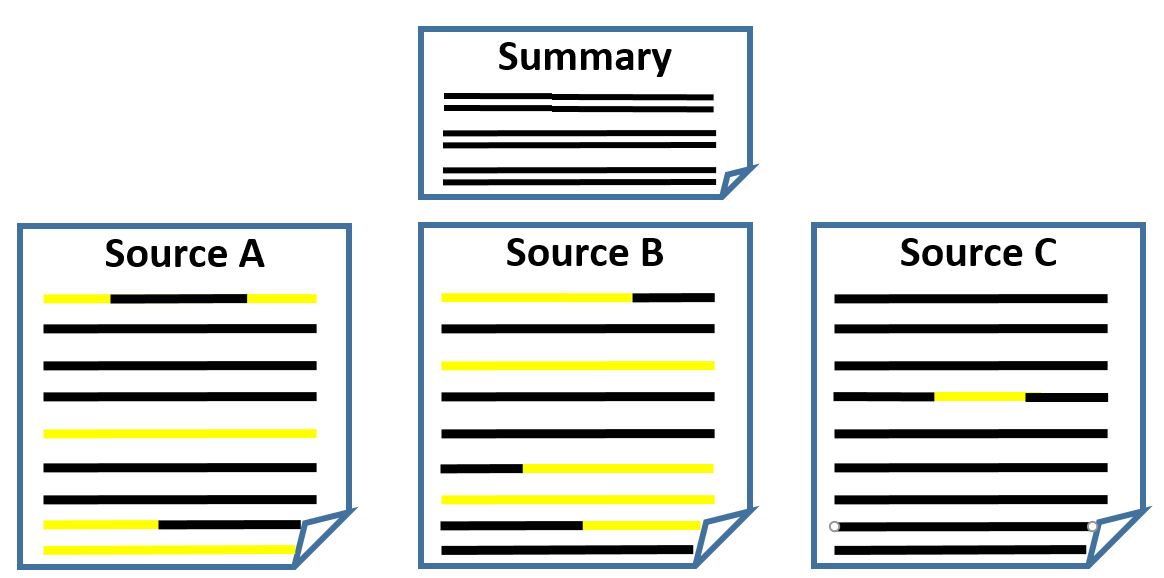}\label{fig_derived_datasets_incontext_fusion}}
\caption{Deriving SPARK task datasets from our alignments, for a given document set (topic): (a) \textbf{Alignments} - aligned summary-source propositions are marked here by the same color;
(b) \textbf{Salience Detection} - all aligned document propositions are to be selected; (c) \textbf{Proposition Clustering} - document propositions aligned with the same summary proposition are to be clustered; (d) \textbf{Evidence Detection} - a summary proposition is the input query, and the document propositions aligned with it are to be extracted as evidence;
(e) \textbf{Text Planning} - document proposition clusters are to be grouped and ordered according to the summary sentence structure; (f) \textbf{Sentence Fusion} - document propositions aligning to the same summary sentence are to be fused to generate that sentence; (g) \textbf{In-context Fusion} - all document propositions, marked within the documents, are to be fused to generate the full summary.}
\label{fig_derived_datasets}
\end{figure*}

Out of the summary-source alignments, annotated manually or automatically, we derive six new datasets for six different tasks, as elaborated below. The tasks are illustrated in \autoref{fig_derived_datasets} and an example topic from our manual dataset is presented in Appendix \ref{app_data_example}. We denote this data suite as ``SPARK'', for Summary Proposition Alignment for Reconstructive Knowledgebases.

\subsection{Salience Detection}
\label{sec_tasks_salience}

Salience detection is the task of marking the important spans within a given source text. It mainly addresses the need within summarization to extract the information around which to summarize the source text \citep{arumae2019highlights}, either extractively \citep[e.g., ][]{mao2020rlmmr} or abstractively \citep[e.g., ][]{chen2018fastabsrl}. Nevertheless, it can be used as a means to merely highlight central parts of a text for easing on a reader \citep{self2013salienceHighlighting, Sandor2014salienceHighlighting, Ponce2022highlighting}.

\paragraph{Task definition and dataset derivation.}
Given document set $D$, the task is to mark the spans in $D$ that globally represent the essential information required to obtain a high level overview of $D$.
From our alignments data, this translates to detecting the spans $H$, i.e., those spans in $D$ that align to the corresponding reference summary $s$ (\autoref{fig_derived_datasets} (b)). Since $s$ is presumably a good portrayal of an overview of $D$, the spans of $H$ should indeed cover the appropriate information.
While the amount and preference of salient information are factors that could be taken into consideration for this task, we rely on the choices of the expert summarizers in the underlying MDS dataset.

From \autoref{tab_dataStat}, we can infer that there are 100 instances for the task (topics) in the test set, and each instance has an average of 22.6 expected spans to identify in the document set.

\paragraph{Evaluation.}
For evaluation we followed \citep{tjong-kim-sang-buchholz-2000-introduction} and used $F_1$ score on the token-level.

\subsection{Proposition Clustering}
\label{sec_tasks_coref_clustering}
Information is expressed differently across sources, and this is especially the case in sets of documents on a related topic, as in our setting. When given a list of propositions, grouping together redundant paraphrastic units is a basic need for gathering and organizing content. 
In summarization and related contexts, redundancy clustering supports generating non-redundant texts that merge overlapping complementary pieces of information. Furthermore, repetition of information typically provides an indication for its importance \citep{wan2008clustering, cai2010clustering, zhang2015clustering}.

In the broad context of paraphrasing, prior datasets generally address paraphrase pairing \citep{dolan2005paraphrases, ganitkevitch2013ppdb}, while our dataset presents the vaster challenge of paraphrase clustering. For \textit{short} text clustering, prior datasets cluster topically related instances, rather than paraphrastic ones \citep{phan2008classifyShortText, xu2017clusteringShortText, cohen2022mcphrasy}. 
Finally, we suggest that paraphrastic matching is better captured at the proposition level, rather than at the sentence level, as a mechanism to prevent misalignment of information.

\paragraph{Task definition and dataset derivation.}
Given a set of proposition-style text units, this task requires producing non-overlapping clusters of units, such that a cluster contains texts that 
express the same meaning or occurrence.
Taken from our alignments data, the text units in a cluster are all the spans across the document set that align to the same proposition in the corresponding reference summary (\autoref{fig_derived_datasets} (c)).

In the test data (\autoref{tab_dataStat}), there are 100 instances (topics) for the task, each with an average of 22.6 spans that need to be clustered into 13.3 clusters. A cluster has an average of 1.7 spans, where 577 clusters are singletons.

\paragraph{Evaluation.}
The traditional clustering metrics are applicable for this task, namely homogeneity (clusters contain only instances that are members of the same gold cluster), completeness (instances that are members of the same gold cluster are also placed together in a predicted cluster), and V-measure (harmonic mean of the first two). In \S{\ref{sec_benchmark}} we only report the V-measure for simplicity.

\subsection{Evidence Detection}
\label{sec_tasks_evidence_detection}
Given a set of documents and a proposition-like phrase, the goal of this task is to find all mentions of the phrase within the documents.
For summarization, this could assist in providing attribution for the summary content \citep{ernst2022procluster, hosking2023attributableSumm, xiao2023salienceExplanationMDS}. It also relates to fact extraction and verification, where a claim needs to be backed by evidence from within a corpus \citep[e.g., ][]{schuster2021factVer}, and coreference search \citep{eirew2022corefSearch} where corefering mentions of an event are to be detected.

\paragraph{Task definition and dataset derivation.}
Given document set $D$ and a textual query $q$, the task is to return all mentions of $q$ within $D$.
With respect to the alignments data, a query is a summary proposition, and the mentions are the document spans that align to it (\autoref{fig_derived_datasets} (d)).

In the test data (\autoref{tab_dataStat}) there are 1332 instances (total number of clusters) for the task. A query requires retrieving an average of 1.7 spans from the document set (with $\sim$3 documents).

\paragraph{Evaluation.}
To evaluate this task we followed the coreference-search evaluation \citep{eirew2022corefSearch} and \citep{tjong-kim-sang-buchholz-2000-introduction}, and used token-based $F_1$.

\subsection{Sentence and Paragraph Planning}
\label{sec_tasks_planning}

To produce a coherent text passage, it is necessary to plan the ordering of the information incorporated into the passage.
This intermediate task was shown to guide models to generate better results \citep{moryossef2019stepbystep}, and has been applied for various generation tasks \citep{barzilay2008localCoherence, chambers2008eventChains, faille2020generationPipeline}. While most related works perform an evaluation extrinsically on the downstream generation task, we establish a dataset explicitly dedicated to ordering and sentence planning.

\paragraph{Task definition and dataset derivation.}
Given a list of proposition clusters $\{C_1, ..., C_k\}$, where a cluster $C_i$ represents a single piece of information,
the task comprises two steps. (1) The clusters need to be ordered so that the respective information flows coherently. (2) After ordering, consecutive clusters that should construct a single sentence are to be grouped. Eventually, this two-stage planning task renders a layout for how to generate a passage containing all the information, in terms of passage- and sentence-level construction. As illustrated in \autoref{fig_derived_datasets} (e), based on the alignments data, each proposition cluster is the set of spans in the document set that align to the same summary span. The ordering and grouping decisions are based on the summary structure: (1) the order of the clusters will be in accordance to the order of respective aligned spans in the summary, and (2) each cluster grouping corresponds to summary spans that come from the same summary sentence.

In the test data (\autoref{tab_dataStat}) there are 100 instances (topics) for the task. An instance has an average of 13.3 information units (clusters) that require planning for passages with 8.3 sentences.

\paragraph{Evaluation.}
To evaluate the ordering of information clusters we used Kendall-Tau correlation between the predicted and the gold ordering (following \citep{lapata-2006-automatic}. For the cluster grouping, we used Homogeneity, Completeness and V-measure, viewed as a clustering assignment.

\subsection{Sentence Fusion}
\label{sec_tasks_sentence_fusion}

Fusing various pieces of information into a single coherent sentence is a fundamental task that is required in many generation tasks, and specifically in summarization where the desired sentence should be concise. Traditionaly, sentence fusion \citep{barzilay2005sentenceFusion} may generally involve two merging scenarios: fusing similar information by omitting redundancy but exploiting complement information from different mentions \citep{thadani2013fusion}, versus combining different information with discourse relation \citep{geva2019discofuse}. These two different fusion types were often addressed separately in prior datasets, while our dataset poses the two challenges simultaneously.

\paragraph{Task definition and dataset derivation.}
Given one or more clusters of paraphrastic texts, the task is to merge all the texts into a single coherent sentence that reflects the union of information in the texts. Furthermore, the information in the generated sentence should generally be presented in the order of the clusters, if more than one is given. The illustration in \autoref{fig_derived_datasets} (f), portraying the derivation of data from alignments, shows that a cluster of texts consists of the spans from the source documents that align to the same summary sentence. Accordingly, the summary sentence acts as the fused sentence. If the sentence consists of more than one proposition, then all corresponding clusters of alignments act as the input.

In the test data (\autoref{tab_dataStat}) there are 834 instances (summary sentences) for the task. Each sentence is a fusion of an average of 2.7 propositions ($\sim$1.6 clusters with $\sim$1.7 propositions).

\paragraph{Evaluation.}
Following \citep{lebanoff-etal-2020-learning, brook-weiss-etal-2021-qa} we apply lexical similarity between the predicted and the gold sentence using ROUGE $F_1$.

\subsection{In-context Fusion}
\label{sec_tasks_incontext_fusion}

Following \citet{slobodkin2022ctr}, another valuable task is generating a passage that consolidates highlights marked within documents. The context around the highlights should assist in resolving anaphora and coreference issues when generating the output. This ability is applicable on its own, e.g., to help a user prepare an abstract out of specially desired content \citep{slobodkin2023summhelper}. Likewise, it can naturally be used in a summarization process where content is first selected, followed by the in-context fusion step to generate the output. The independent fuser allows for any conditional selection of highlights (e.g., for query-focused summarization). Our dataset is different from that of \citet{slobodkin2022ctr} in that we address the multi-document setting, where the input is larger and redundancy is a more prevalent phenomenon.

\paragraph{Task definition and dataset derivation.}
Given a set of documents and marked spans within the documents, the task is to generate a coherent passage that contains all and only the information in the marked spans. With respect to our alignments data, the highlights are all the spans aligning to the reference summary, and the summary acts as the fused passage to generate (\autoref{fig_derived_datasets} (g)).

In the test data (\autoref{tab_dataStat}) there are 100 instances (topic) for the task. Each document set consists of an average of 22.6 spans that need to be fused into a passage.

\paragraph{Evaluation.}
We used ROUGE $F_1$ between the predicted and the gold passage.

\section{Baseline Experiments}
\label{sec_benchmark}
We next examine the performance of current technology on the six aforementioned datasets. For each task, we consider a dedicated trained model and an execution of \texttt{gpt-3.5-turbo}, once in zero-shot mode and once with an in-context example (prompts in Appendix \ref{app_sec_prompts}). Since the large size of a multi-document set limits model architecture options, we resorted to models devised for the multi-document setting. We used our train set to train the dedicated models (\S{\ref{sec_dataCollection_auto}}).
In addition, the large input sizes afforded us to execute GPT with only one in-context example.
We explain the finetuned models in \S{\ref{sec_baslines_models}} and discuss the results in \S{\ref{sec_baselines_results}}.

\subsection{Finetuned Models}
\label{sec_baslines_models}

For \textbf{Salience Detection}, we finetuned the Cross-Document Language Model \citep[\texttt{CDLM}; ][]{caciularu2021cdlm}, an encoder-only model that was trained specifically to handle multi-document inputs by assigning global attention from selected tokens to the entire document set. In our case, we added a classification head and input the document set with special tokens marking a candidate span, while the target is a binary decision for whether a span is salient or not. Global attention was assigned to the candidate tokens. At inference time, we must mark candidate spans within the document set for the model to classify for salience. To that end we use Open Information Extraction \cite{stanovsky2018oie}, following \cite[][which was also used to create the train set]{ernst2021superpal}.

For both the \textbf{Proposition Clustering} and \textbf{Evidence Detection} tasks we employ the \texttt{SuperPAL} \citep{ernst2021superpal} model which was pre-trained to match pairs of similar spans. For each pair, the model outputs a score between 0 (no match) and 1 (match). For Proposition Clustering, the scores were used as input for an Agglomorative Clustering method to group similar spans. For Evidence Detection, we paired each summary span with each candidate \textit{document} span, and selected spans scored above the original SuperPAL threshold (0.5). 

For the \textbf{Sentence and Paragraph Planning} and \textbf{Sentence Fusion} tasks, we finetuned \texttt{Flan-T5-XXL} \citep{chung2022flant5} as a sequence-to-sequence task with suitable instructions for each of the tasks (detailed in Appendix \ref{app_sec_prompts}). For the Planning task, we grouped together all document spans that are aligned to the same summary span, selected a random representative per group, and numbered each representative. Then the model is required to output an ordered list of list of indices, where each sub-list represents a prospective summary sentence, and the order of the sub-lists outlines the passage structure. For Sentence Fusion, the model recieves a group of document spans and is expected to generate the aligned summary span.

For the \textbf{In-context Fusion} task, we used the \texttt{QAMDen} model \citep{caciularu-etal-2023-peek}, a recent encoder-decoder transformer made for the multi-document setting, and pre-trained on Multi-News. We finetuned the model for our task by surrounding the highlighted spans with special tokens, akin to \citet{slobodkin2022ctr, slobodkin-etal-2023-dont}.

\begin{table*}[!tbp]
\centering
    \resizebox{\textwidth}{!}{
    \begin{tabular}{l|c|c|c|c|c|c|c|c|c|c|c}
    \toprule
       
        \textbf{Task} & \textbf{Salience} & \textbf{Clustering} & \textbf{Evidence} & \multicolumn{2}{c|}{\textbf{Planning} } & \multicolumn{3}{c|}{\textbf{Sent. Fusion} }& \multicolumn{3}{c}{\textbf{In-context Fusion}} \\
        \textbf{Metric} & $F_1$ & V& $F_1$ & Kendall's $\tau$& V &  R-1 & R-2 & R-L & R-1 & R-2 & R-L \\
        \midrule
        Finetuned & 0.49 & 0.33 & 0.36 & 0.36 & 0.76 & 0.45 &0.26 & 0.39 & 40.54 & 16.82 & 22.42\\ 
        GPT Zero-shot & 0.27 & 0.71 & 0.22 & 0.29 &  0.70 &0.43 & 0.22 & 0.34& 38.45 & 13.29 & 19.94 \\
        GPT In-Context & 0.31 & 0.83 & 0.32 & 0.33& 0.67 & 0.38& 0.17& 0.29 & 40.01 & 13.65 & 20.43\\
        \bottomrule

    \end{tabular}}
    \caption{Performance of finetuned, zero-shot GPT, and in-context learning GPT models on all tasks. Overall, a smaller finetuned model yields better results than the GPT counterpart.}
    \label{tab_main_res}
    
\end{table*}

\subsection{Results}
\label{sec_baselines_results}

Results are presented in \autoref{tab_main_res}. As can be seen, even though we used a much larger model for the zero-shot and in-context modes (\texttt{gpt-3.5-turbo}), the finetuned models perform better in all tasks except for Proposition Clustering. Apparently, when the input data is 
short,
without having to input all documents, the GPT model performs better. In addition, the score differences between the models are quite low on the two Fusion tasks, where the output performance is more subjective. We find that the in-context example assists GPT in most tasks with respect to zero-shot mode. The sentence fusion was harmed since GPT tended to ask for additional details in its output, as it tried to comply with certain characteristics of the example it received in-context.
Overall, future research can examine how to push ahead strong large language models either with further fine-tuning or with in-context examples of very large inputs.

\section{Source Dataset Characteristics}
\label{sec_analysis}

\begin{table}[!tbp]
\centering
    \resizebox{\linewidth}{!}{
    \begin{tabular}{lrrr}
    \toprule
        \textbf{Overlap Measure} & \textbf{unigram} & \textbf{bigram} & \textbf{trigram}  \\
    \midrule

    \textit{Alignment Pair} & 43.39 & 23.44 & 16.12   \\
      \textit{Cluster Max} & 51.81 & 31.38 & 22.92   \\
      \textit{Full Cluster} & 54.20  & 32.26 & 23.24  \\
      \textit{In-Cluster} & 35.72  & 17.60 & 11.64  \\
      
   \bottomrule


    \end{tabular}}
    \caption{Percentage of n-gram overlap of different source span groups with respect to their aligned summary span. Overall, summary spans are partially abstractive. We also measure the n-gram overlap between document spans within the same cluster (\textit{'In-Cluster'}). This indicates lexical diversity in redundant source spans.}
    \label{tab_abstractiveness}
    
\end{table}

The alignments extracted from the MDS dataset (Multi-News) sheds light on various characteristics of the source dataset. It helps understand the amount of information redundancy, level of abstractivness, and spread of content within documents.

\paragraph{Information redundancy.}
From the alignment data statistics in \autoref{tab_dataStat}, the average size of clusters is 1.7 propositions from 3 documents, which reflects on the low informational redundancy within a document set. Assuming information redundancy impacts apparent importance, this property may affect the ability to recognize salient information and plan passage structure.

\paragraph{Abstractiveness.}
A cluster of document-set spans and its respective summary span can be examined for paraphrastic differences to measure abstractiveness within the data. To that end we present in Table \ref{tab_abstractiveness} the conventional n-gram overlap metric between spans on several levels: (1) \textit{Alignment Pair} is the percentage of summary span n-grams that appear also in the document span; (2) \textit{Cluster Max} is the maximum pair overlap score in a cluster, which indicates general summary-source abstractivness; (3) \textit{Full Cluster} is the n-gram overlap between the bag of document spans in a cluster with respect to their aligned summary span; (4) \textit{In-Cluster} is the average pairwise overlap between cluster members. \textit{Full Cluster} is only slightly higher than \textit{Cluster Max}, as additional members of the cluster do not contribute much to cover the summary span. This indicates that the summary span is mostly copied from a single document span, and does not merge texts from different places in the documents. To strengthen this insight, \textit{In-Cluster} produces a relatively low score, meaning that abstractiveness is high within a cluster, while one of the cluster members is more lexically similar to the summary span. We can also learn that the relatively low number of clusters per summary sentence (1.6) indicates that summaries only require occasional fusion of different information units. Overall, these insights reinforce that the summaries in the Multi-News MDS dataset have somewhat low abstractiveness, as also observed from \citet{fabbri-etal-2019-multi}, though when information repeats within a document set, it is mentioned in noticeably different phrasing.

\paragraph{Spread within Documents.}
Since a summary is comprehensively aligned with its corresponding document set through our data, we can investigate the individual importance of each document for the summary content \citep{wolhandler2022howmulti}. We find that out of the $\sim$3 documents per topic, only $\sim$87\% of documents have aligning information with the summary on average, meaning some topics do not require all the documents for the summary. On the cluster level, we find that a summary span aligns only to $\sim$53\% of the documents on average. This stresses the low information redundancy mentioned before, where each summary proposition appears in one or two documents.


\section{Conclusion}
\label{sec_conclusion}
We advocate the potential utility of proposition-level summary-source information alignment, particularly in the multi-document setting, for exposing a wide range of summarization-related tasks. Specifically, we reveal that these alignments induce datasets for a broad range of appealing tasks arising in summarization, and applicable as standalone tasks. We annotated a high-quality test dataset of alignments, and automatically compiled large-scale train and dev sets. From the alignments data, we automatically derived datasets for six distinct tasks. Our released dataset collection, along with our baselines and analyses, promotes future research on a challenging multi-text task suite.


\section*{Limitations}
\label{sec_ethical}
This study obtains alignments out of a MDS dataset in the news domain. To automatically extract alignments, we leveraged SuperPAL, which itself is trained on news data. The model would likely extract less accurate alignments in other domains. Generally, it is worthwhile to perform an alignment study like ours in additional domains and languages. The alignment process and guidelines, as well as the derived tasks may differ in accordance to the source MDS data.

The quality of our alignments is dependent on the quality of the source MDS dataset (Multi-News) from which we extract the alignments. For example if a reference summary is not fully faithful or comprehensive for some reason, this may have an effect on our alignment assumptions. Our analysis in \S{\ref{sec_analysis}} sheds light on some of these discrepancies.

The baselines we presented are limited to the prompts we used. Other prompts may yield different results.

\bibliography{bibliography}

\appendix
\section{Further Details on Baseline Implementations}
We describe here details regarding the baseline models outlined in \S{\ref{sec_baslines_models}}. Specifically, for each task, we describe heuristics applied in case an LLM model outputs an answer in the wrong format (relevant for the GPT and finetuned Flan baselines).

\paragraph{Salience Detection and Evidence Detection.}
In these tasks we asked the model to extract (fully copy) spans from the source text. However, in many cases the model slightly changed the extracted span by adding or omitting a word. Since our evaluation for these two tasks is token-level $F_1$, we need to locate the extracted spans in the source documents. To do so, we extracted proposition candidates \citep[using OpenIE; ][]{stanovsky2018oie} from the source, and for each predicted span, we found the OpenIE proposition with the highest lexical overlap.

\paragraph{Proposition Clustering.}
In this task we asked the model to cluster spans by predicting a cluster index for each span. However, in some of the cases the model did not provide an index for an input spans. In such cases, we assign a random (existing) cluster index to this text span.

\paragraph{Sentence and Paragraph Planning.}
The model is tasked with outputting a list of lists describing the order of information (span clusters) within the final paragraph, each represented by its index. 
In some cases, the model omits an index of one of the spans, adds a non-existent index, or even repeats an existing index more than once. To cope with this, we removed non-existing indices, kept only the first occurrence of a repeating index, and appended a randomly ordered list of missing indices.

\paragraph{Sentence Fusion and In-Context Fusion.}
As these two tasks generate free text and are evaluated by ROUGE, mis-formatting is not relevant.

\section{Model Prompts}
\label{app_sec_prompts}
\autoref{table:gpt-prompts} presents the prompts used on the \texttt{gpt-3.5-turbo-0301} model for the six tasks, in zero-shot and in-context learning modes. \autoref{table:non-gpt-prompts} shows the prompts used for finetuning a \texttt{flan-t5-xxl} model for the Planning and Sentence Fusion tasks.

\begin{table*}[t]
    \centering
    \begin{tabular}{|p{0.2\linewidth}|p{0.75\linewidth}|}
\hline
\textbf{Task} & \textbf{Prompt} \\ \hline

Salience Detection & \texttt{Below are documents on the same topic in different user messages. Please copy exactly salient sub-sentenial spans. Do not change the copied text.} \\ \hline

Proposition Clustering & \texttt{Below are text spans with indexes. Please cluster them into groups. \newline
Each group should contain spans that share the same information. \newline
Return a dict in the following format {<SPAN IDX>: <CLUSTER IDX>}. Do not add anything beside the dict.} \\ \hline

Evidence Detection & \texttt{Below are documents on the same topic and a query. \newline
Please extract exactly short text spans from the documents that match the information in the query. \newline
Separate the spans with a new line and start each span with -.} \\ \hline

Sentence and Passage Planning & \texttt{Your task is to structure a set of information units, each pertinent to a central topic, into a cohesive paragraph.\newline
Begin by analyzing and logically arranging these units to ensure a seamless progression of ideas. \newline
Once you've established a coherent sequence, segment the units into subgroups that represent distinct conceptual sentences. \newline
Your final output should adhere to a Python list of lists format. Each internal list must encompass the indices of information units that belong to a particular conceptual sentence within the paragraph. \newline
Output Examples: \newline
"[[3, 4, 1], [0, 2], [5]]"\newline
Your output format MUST be a simple Python list of lists only, with no comments.
} \\ 
\hline

Sentence Fusion & \texttt{Merge the following text clusters into a single coherent sentence: \newline
Your response MUST contain only one sentence.} \\ 

\hline

In-Context Fusion & \texttt{Below are documents on the same topic. Please summarize the spans marked in <> while using the set of documents as context.} \\ \hline
\end{tabular}
\caption{Prompts used as input to \texttt{gpt-3.5-turbo} to solve each of the six tasks in zero-shot mode. Similar prompts were used for the in-context learning mode, with the addition of an example.}
\label{table:gpt-prompts}
\end{table*}

\begin{table*}[t]
    \centering
    \begin{tabular}{|p{0.2\linewidth}|p{0.75\linewidth}|}
\hline
\textbf{Task} & \textbf{Prompt} \\ \hline

Sentence and Passage Planning & \texttt{Task: Paragraph planning \newline
Your task is to structure a set of information units, each pertinent to a central topic, into a cohesive paragraph.\newline
Begin by analyzing and logically arranging these units to ensure a seamless progression of ideas. \newline
Once you've established a coherent sequence, segment the units into subgroups that represent distinct conceptual sentences. \newline
Your final output should adhere to a Python list of lists format. Each internal list must encompass the indices of information units that belong to a particular conceptual sentence within the paragraph. \newline
Output Examples: \newline
"[[3, 4, 1], [0, 2], [5]]"\newline
} \\ 
\hline

Sentence Fusion & \texttt{Merge the following text clusters into a single coherent sentence:\newline
} \\ 
\hline
\end{tabular}
\caption{Prompts used as input to \texttt{flan-t5-xxl} for finetuning the model for the respective tasks.}
\label{table:non-gpt-prompts}
\end{table*}

\section{Full Annotation Guidelines}
\label{appendix_annotation_guidelines}

\begin{figure*}[ht!]
\centering
    \includegraphics[width=16cm]{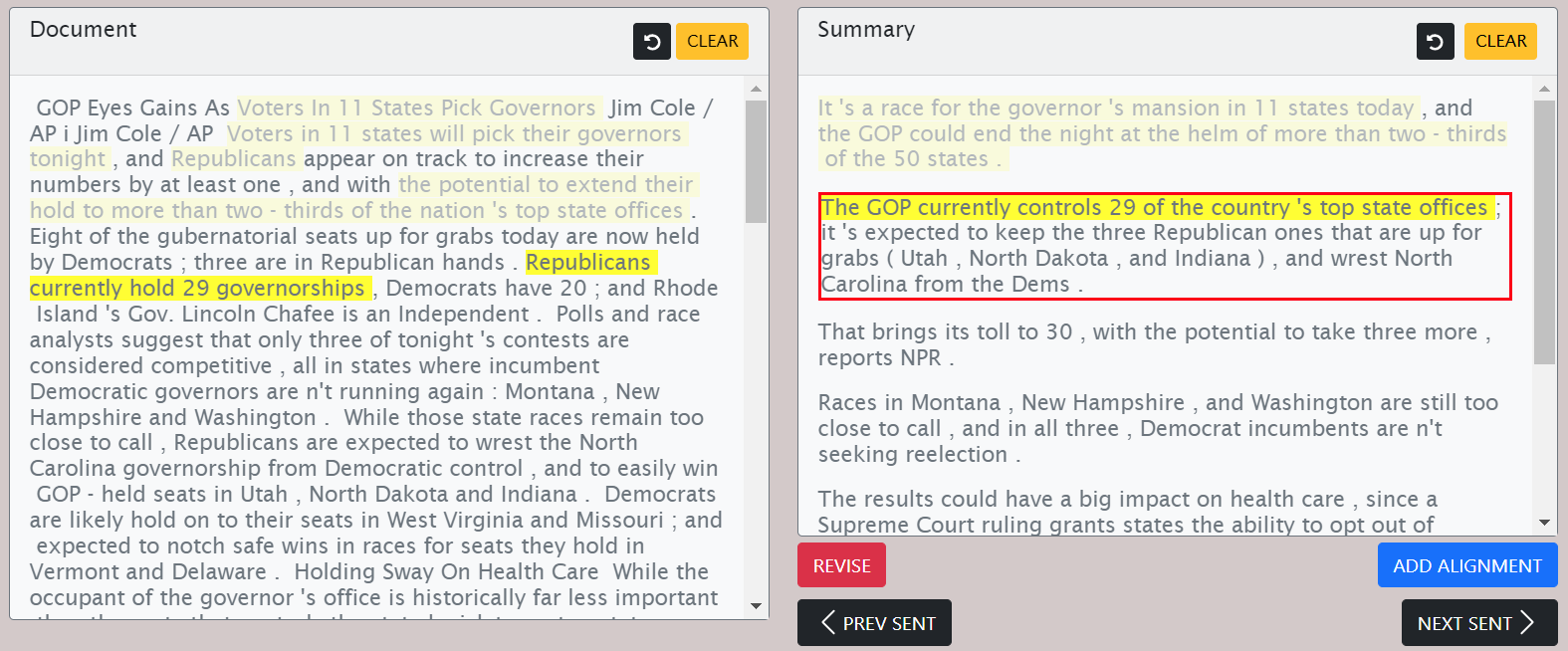}
    \caption{The alignment annotation interface. The annotator marks a span (proposition) in the summary (right) along with all matching spans in the current document (left). To minimize cognitive load, a summary is shown next to a single document at a time, and the procedure is conducted separately for all documents in the document set. Also visual focus is placed on one summary sentence at a time (red rectangle) to orient the process.}
    \label{fig:UI_example}
\end{figure*}
This section describes the complete annotation guidelines for the crowdsourced alignment procedure.

\subsection{Summary-related Guidelines}\label{subsec:Summary-related Guidelines}
As mentioned in \S\ref{subsec:guidelines}, we guide annotators to separate summary sentences into separate events, focusing on one event at a time.
An event is identified as a predicate alongside all its arguments, with instructions for annotators to include all associated arguments, even if repeated across events, e.g., \textit{``Jane came by and left''}, where \textit{``Jane''} is part of both the \textit{``came by''} and \textit{``left''} events.

We also aim to address facts represented in various grammatical forms, but for the sake of simplification for the annotators, we highlight the following two forms:
$\bullet$ \textsc{Secondary Verb}: This involves nested events, where a smaller event serves as an argument for a larger one, e.g., \textit{``John insisted on inviting her''}. Annotators are guided to merge these into a single event if both appear in the source document, but to align only the nested event if it alone is present. 
Annotators are also guided to distinguish between prospective and transpired events. For instance, \textit{``John insisted on inviting her''} (prospective) should not align with \textit{``John invited her''} (transpired) since they convey different events. Moreover, in instances of nested spans containing distinct events, like \textit{``She said she arrived and went to bed''} annotators should align the primary event with each nested event separately if both are documented.
$\bullet$ \textsc{Connecting Words}:  For events linked by discourse markers, which we refer to in our guidelines as \textit{connecting words}, annotators are trained to identify when these words indicate a genuine connection, such as \textit{``He ate because he was hungry''} versus when they merely place events side by side, like in \textit{``He went home and ate an apple''}.  In the former, both events are to be combined into a single alignment if present in the document, while in the latter, events should be aligned separately.

\subsection{Document-related Guidelines}\label{subsec:Document-related Guidelines}
On the document side, we provide the annotators with the following guidelines on how to align a span to a summary proposition.
$\bullet$ \textsc{Paraphrasing}: We guide our workers to not depend exclusively on phrases with common words, since the matching document phrases are frequently a paraphrase of their summary counterparts.
$\bullet$ \textsc{Consecutiveness}: We instruct our workers to avoid highlighting unnecessary details, and keep the highlights non-consecutive if necessary.
$\bullet$ \textsc{Entailment}: As described in \S\ref{sec_dataCollection}, we instruct the annotators to also align document spans that either entail the summary event, or are entailed by it. E.g., \textit{``John ate an apple''} versus \textit{``John ate fruit''}.
$\bullet$ \textsc{Missing Details}: In cases where some details of the summary event are missing from the currently inspected document, we guide our annotators to leave those un-highlighted on the summary side, and align only the details that do appear.
$\bullet$ \textsc{Exhaustiveness}: We also train our workers to identify all document mentions of the current summary event, and align each one separately.

\section{Obtaining Gold Alignment Clusters}
\label{app_grouping_alignments}
Since the alignment annotation is conducted one document at a time, the coreferring propositions from across documents (those aligning to the same summary proposition) need to be clustered together. Considering that a summary proposition may be marked slightly differently each time (with different boundaries), we allow a 0.5 (tuned threshold) intersection-over-union (of tokens) to consider summary spans as referring to the same proposition. To validate this threshold, we manually examined 10 topics that contain 94 clusters. We found that only one cluster merged irrelevant propositions, and only 3 pairs of clusters should have been merged to a larger cluster. Accordingly, this enables almost perfect clustering of document spans for our data. For the Evidence Detection task, we aggregated the cluster query as the union of all aligned summary spans in this cluster.

\section{Model Training}
\paragraph{Salience Detection. } We trained the CDLM model for 2 epochs with learning rate of 1e-5 and batch size of 3 instances on 3 A100 GPUs for one hour (meaning an effective batch size was 9).

\paragraph{In-Context Fusion. } We trained the PEEK model for 50,000 steps with learning rate of 3e-5 and batch size of 16 instances on 1 A100 GPUs for one hour.

\paragraph{Sentence Fusion and Planning. } We finetuned a Flan-T5-XXL as a Sequence-to-Sequence task using LoRA and applied 8-bit quantization for optimization. We trained using a sample of 10K examples derived from our train-set, for one epoch using a learning rate of 5e-5 and using AdamW optimizer.

\section{Data Example}
\label{app_data_example}
\begin{figure*}[ht!]
\centering
    \includegraphics[scale=0.67]{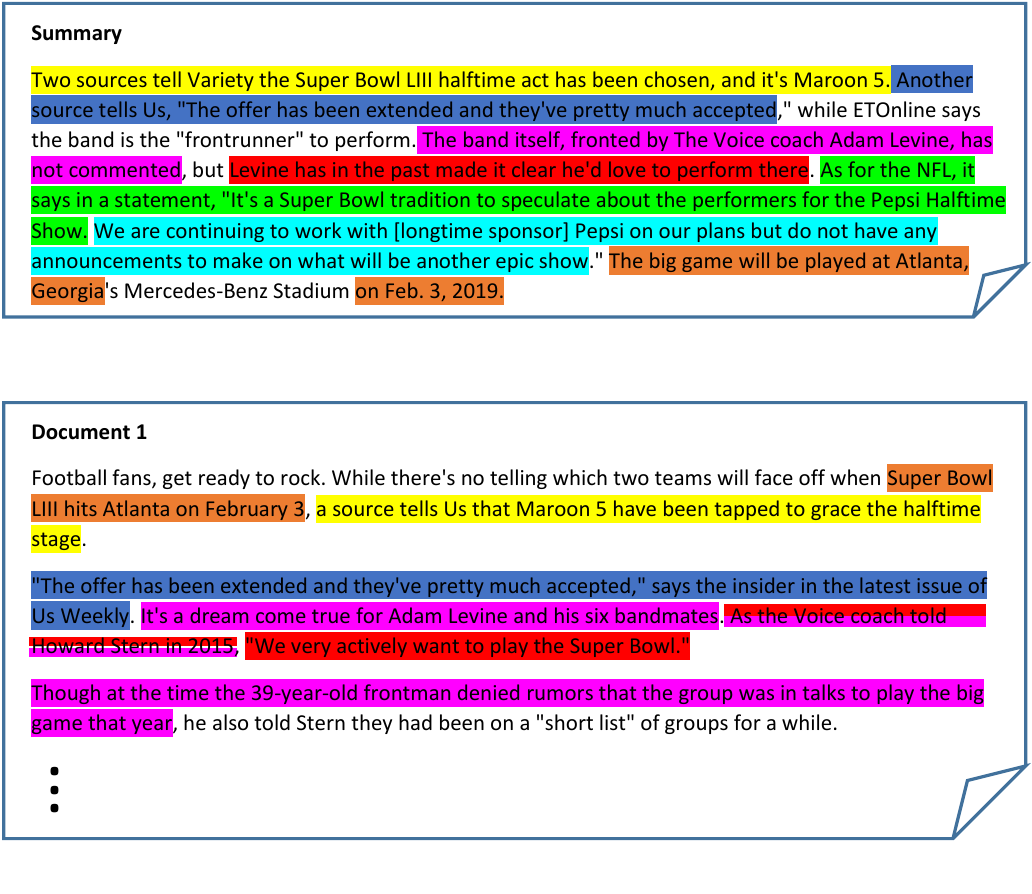}
    \includegraphics[scale=0.67]{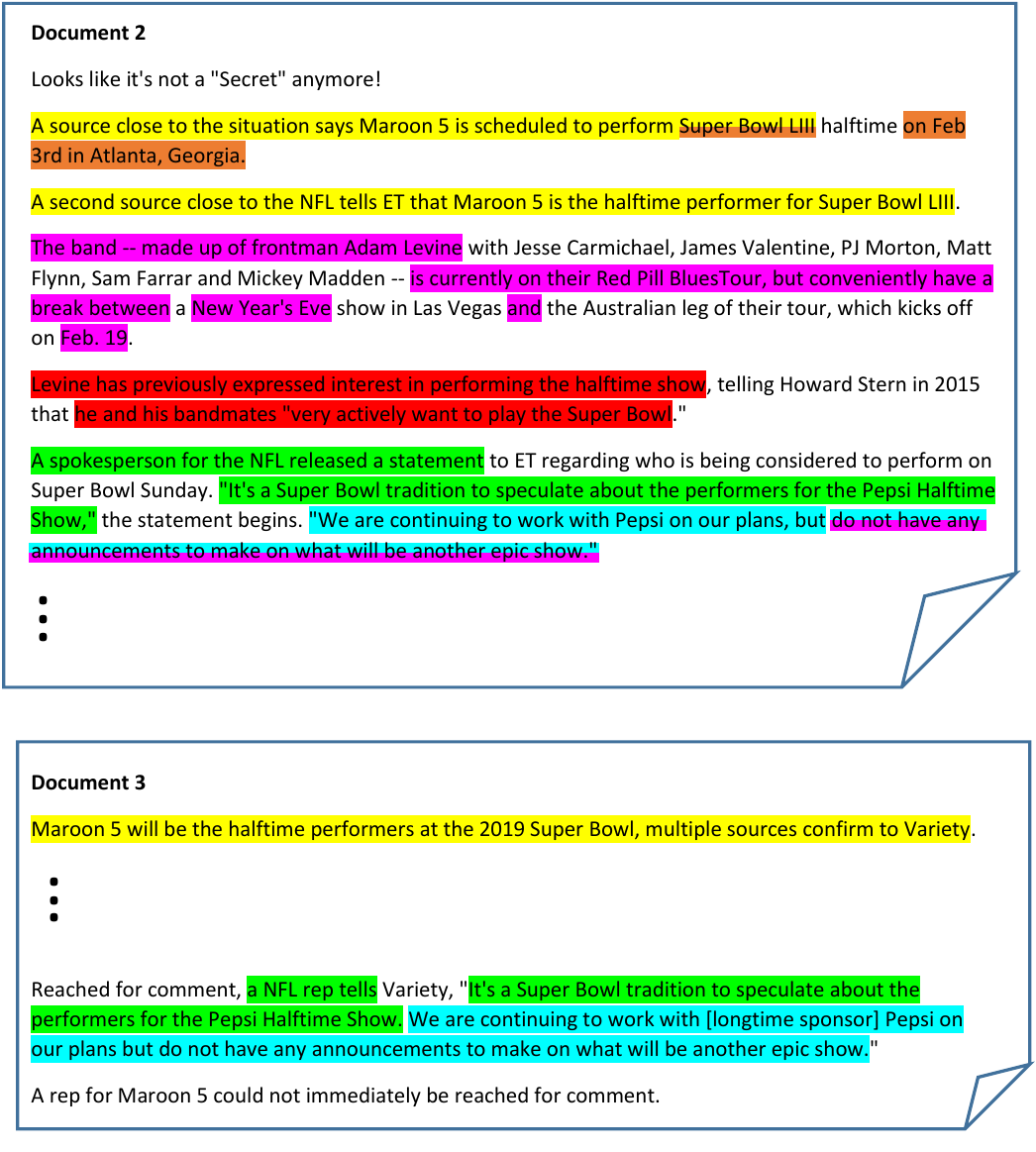}
    \caption{The manual alignment annotation on topic31 from our data. The documents have been shortened for presentation purposes.}
    \label{fig:alignment_text_example}
\end{figure*}

\begin{figure*}[ht!]
\centering
    \includegraphics[scale=0.45]{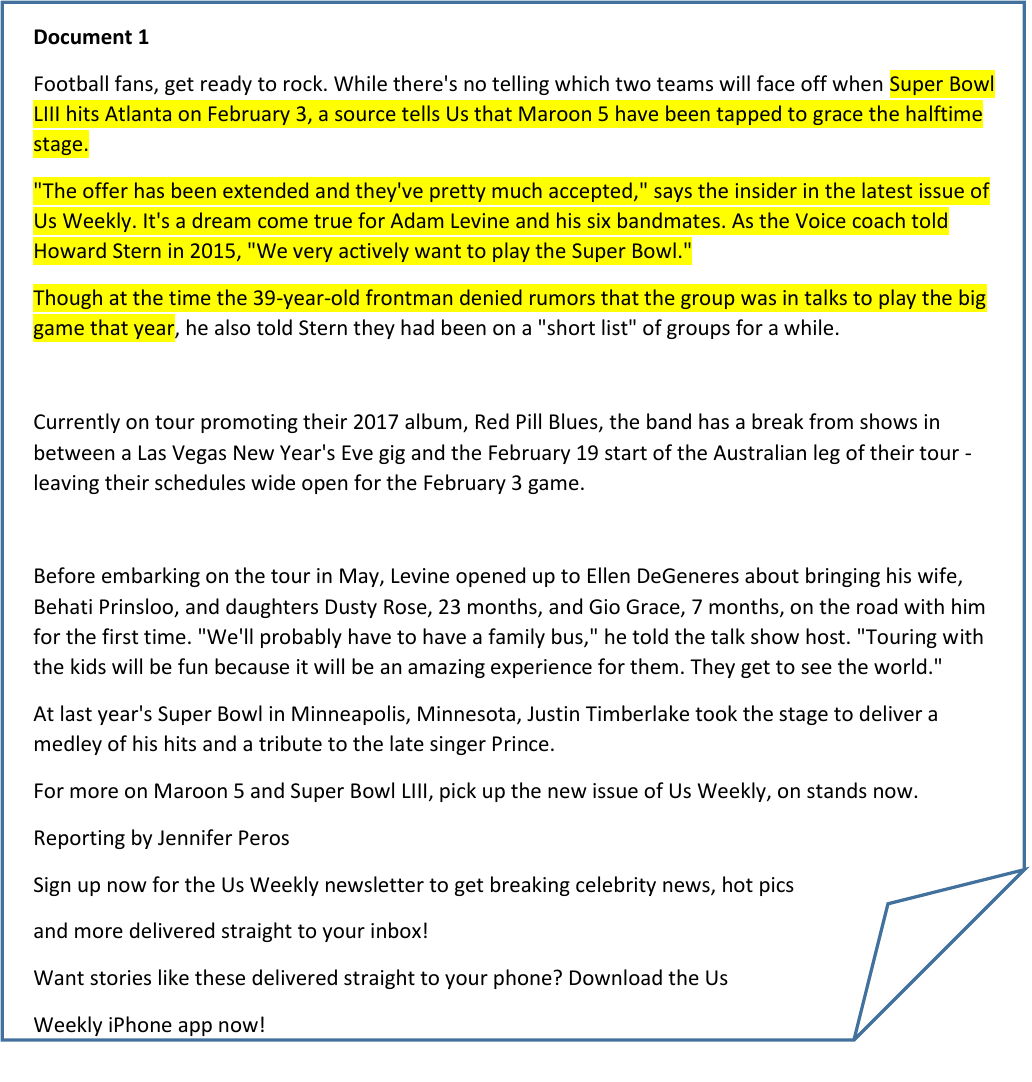}
    \includegraphics[scale=0.45]{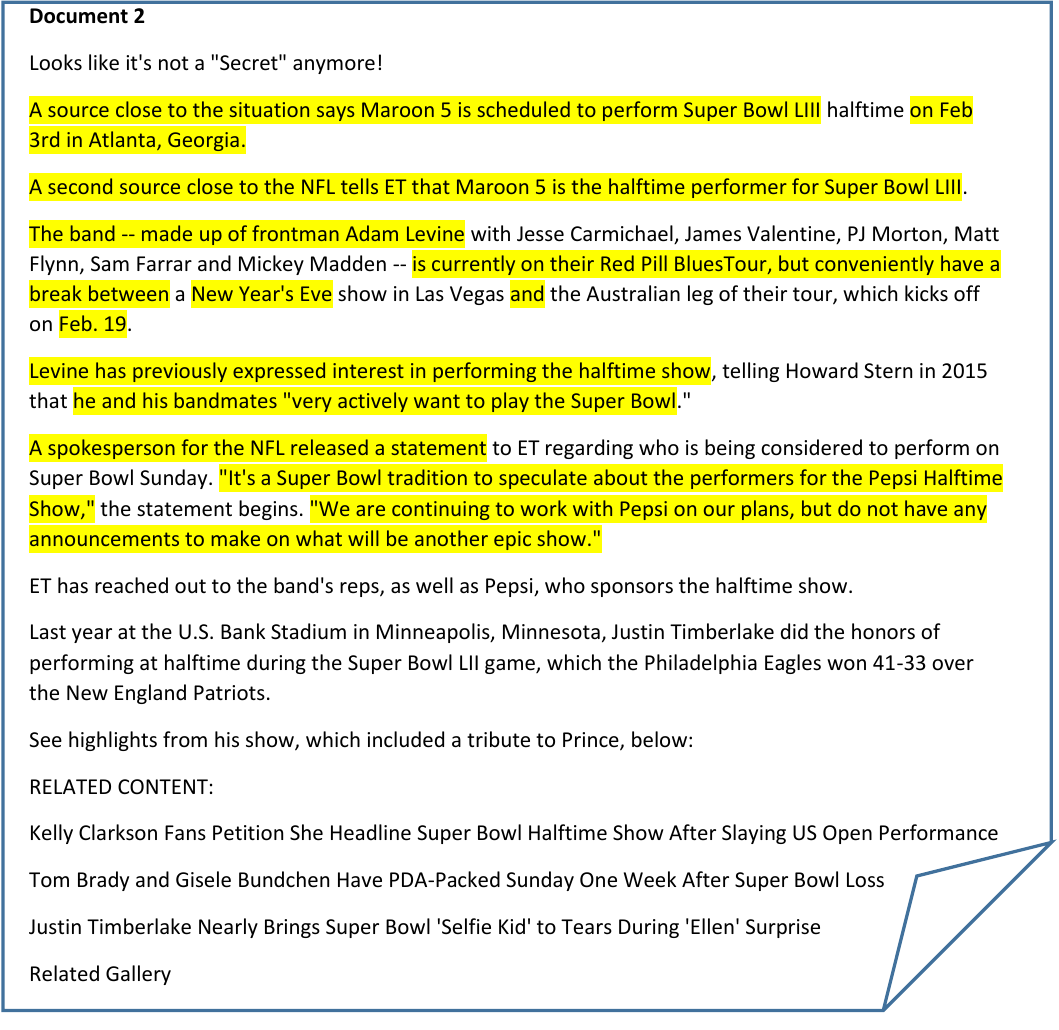}
     \includegraphics[scale=0.45]{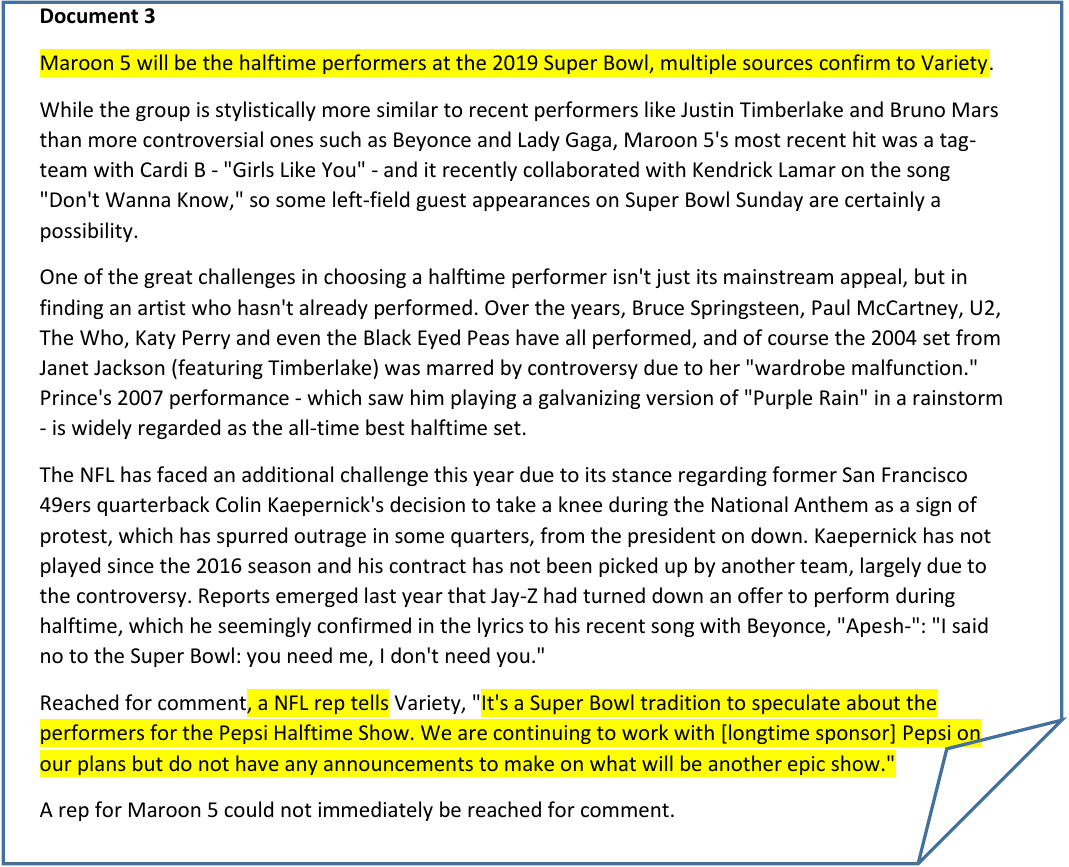}

    \caption{An example of a \textbf{Salience Detection} instance derived from the alignments in Figure \ref{fig:alignment_text_example}. All aligned document propositions are salient. These highlighted documents can also serve as input to the \textbf{In-context Passage Fusion} task, where the output would be the original reference summary. }

    \label{fig:salience_text_example}
\end{figure*}

\begin{figure*}[ht!]
\centering
    \includegraphics[scale=0.8]{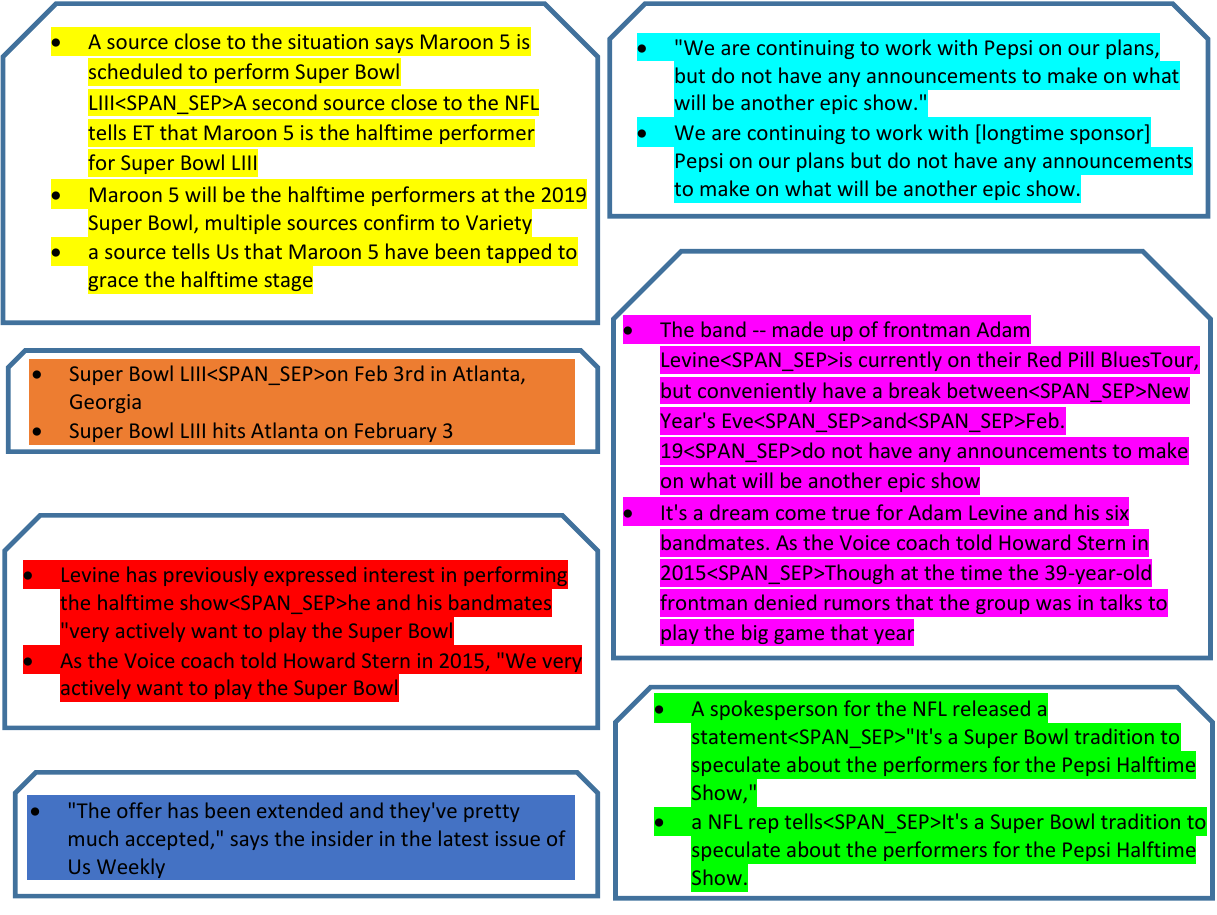}

    \caption{An example of a \textbf{Proposition Clustering} instance derived from the alignments in Figure \ref{fig:alignment_text_example}. Clusters contain document propositions that are aligned to the same summary proposition.}

    \label{fig:proposition_clustering_text_example}
\end{figure*}

\begin{figure*}[ht!]
\centering
    \includegraphics{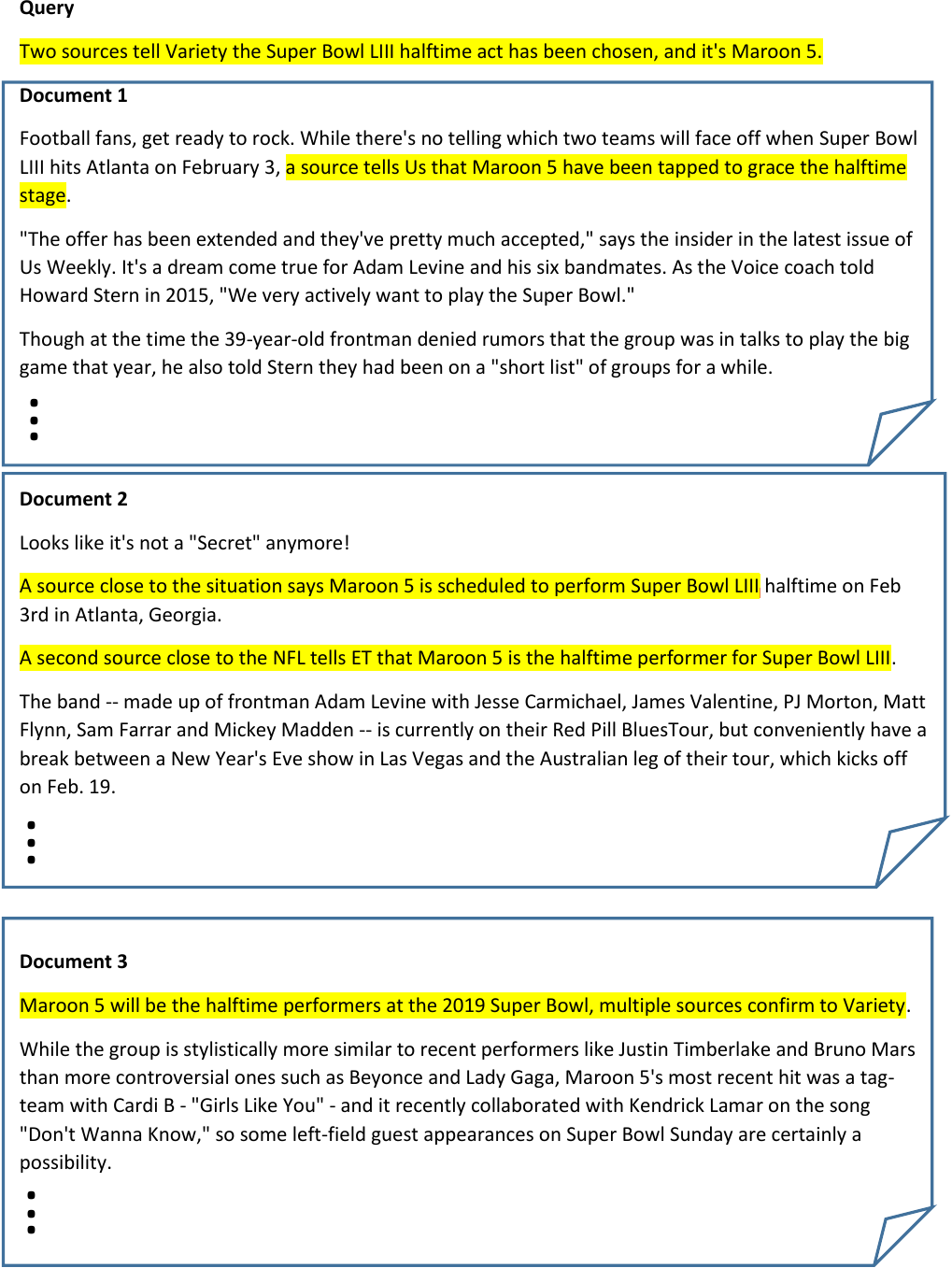}

    \caption{An example of a \textbf{Evidence Detection} instance derived from the alignments in Figure \ref{fig:alignment_text_example}. The evidences are the document propositions aligned to the query from the summary. The documents have been shortened for presentation purposes.}

    \label{fig:evidence_detection_text_example}
\end{figure*}

\begin{figure*}[ht!]
\centering
    \includegraphics[scale=0.8]{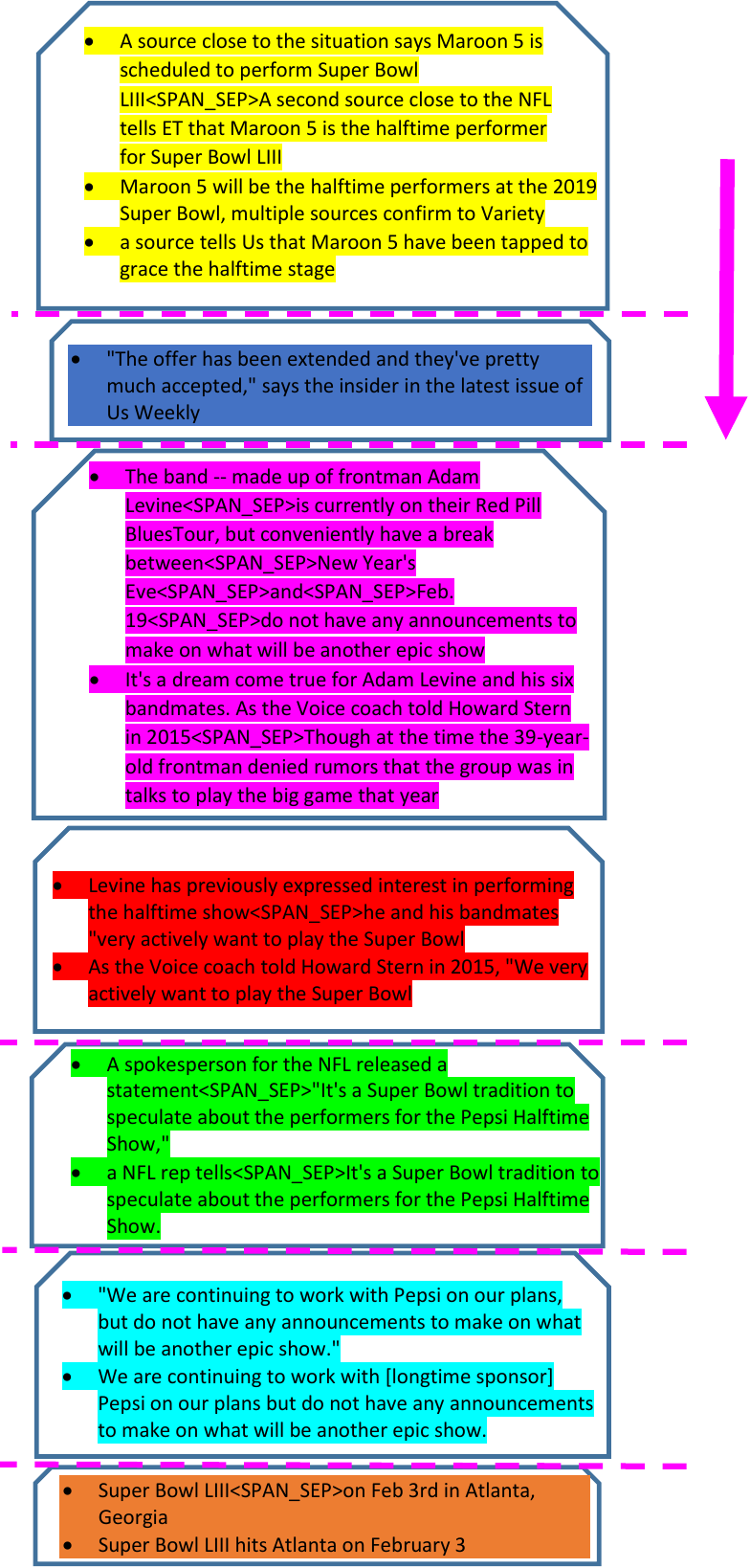}

    \caption{An example of a \textbf{Sentence \& Paragraph Planning} instance derived from the alignments in Figure \ref{fig:alignment_text_example}. The clusters are ordered by the order of their aligned summary propositions, and grouped with clusters with the same aligned summary sentence.}

    \label{fig:planning_text_example}
\end{figure*}

\begin{figure*}[ht!]
\centering
    \includegraphics[scale=0.8]{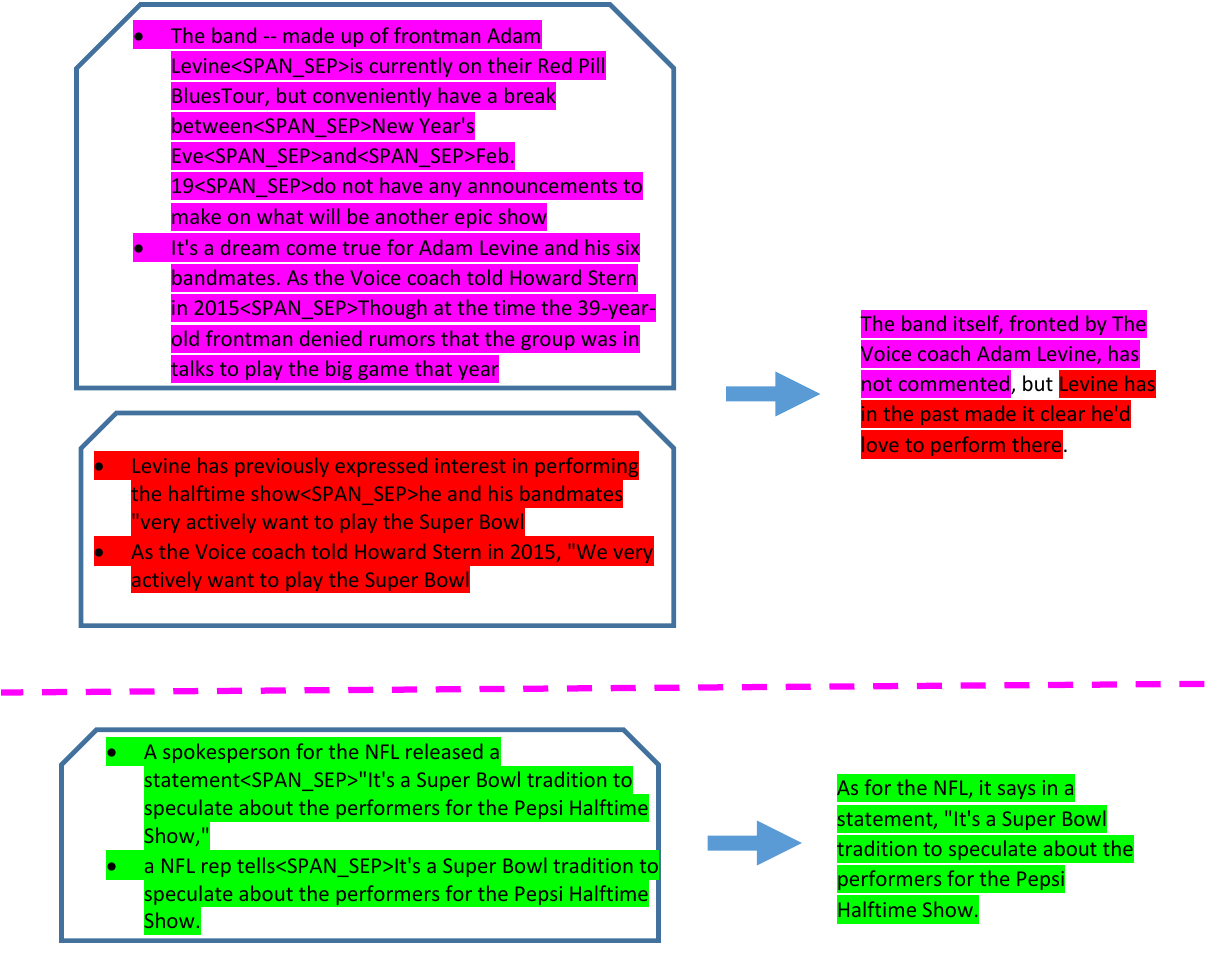}

    \caption{An example of some of the \textbf{Sentence Fusion} instances derived from the alignments in Figure \ref{fig:alignment_text_example}. The clusters that are aligned to the same summary sentence should be fused to generate this sentence.}

    \label{fig:fusion_text_example}
\end{figure*}

We show an example of one topic of our SPARK data suite, starting from the alignment annotation (Figure \ref{fig:alignment_text_example}) followed by its derived instances (Figures \ref{fig:salience_text_example}, \ref{fig:proposition_clustering_text_example},
\ref{fig:evidence_detection_text_example}, \ref{fig:planning_text_example}, \ref{fig:fusion_text_example}). 


\end{document}